\documentclass[10pt,twocolumn,letterpaper]{article}

\usepackage{iccv}
\usepackage{times}
\usepackage{epsfig}
\usepackage{graphicx}
\usepackage{float}
\usepackage{subfig}
\usepackage{amsmath}
\usepackage{amssymb}
\usepackage{booktabs}
\usepackage{multirow}
\usepackage[pagebackref=true,breaklinks=true,letterpaper=true,colorlinks,bookmarks=false]{hyperref}

\iccvfinalcopy % *** Uncomment this line for the final submission
\usepackage[capitalise,noabbrev]{cleveref}

\def\iccvPaperID{7} % *** Enter the ICCV Paper ID here

\ificcvfinal\fi

\begin{document}

%%%%%%%%% TITLE
\title{
DFM-X:  Augmentation by Leveraging Prior Knowledge of Shortcut Learning}

\author{Shunxin Wang
\and
Christoph Brune
\and
Raymond Veldhuis
\and 
Nicola Strisciuglio\and
University of Twente, The Netherlands
}

\maketitle
% Remove page # from the first page of camera-ready.
\ificcvfinal\thispagestyle{empty}\fi

%%%%%%%%% ABSTRACT
\begin{abstract}
Neural networks are prone to learn easy solutions from superficial statistics in the data, namely shortcut learning, which impairs generalization and robustness of models. We propose a data augmentation strategy, named DFM-X, that leverages knowledge about frequency shortcuts, encoded in \textbf{D}ominant \textbf{F}requencies \textbf{Maps} computed for image classification models. We randomly select \textbf{X}\% training images of certain classes for augmentation, and process them by retaining the frequencies included in the DFMs of other classes. 
This strategy compels the models to leverage a broader range of frequencies for classification, rather than relying on specific frequency sets. Thus, the models learn more deep and task-related semantics compared to their counterpart trained with standard setups. Unlike other commonly used augmentation techniques which focus on increasing the visual variations of training data, our method targets exploiting the original data efficiently, by distilling prior knowledge about destructive learning behavior of models from data. Our experimental results demonstrate that DFM-X improves robustness against common corruptions and adversarial attacks. It can be seamlessly integrated with other augmentation techniques to further enhance the robustness of models. Codes are available at \url{https://github.com/nis-research/dfmX-augmentation}.

\end{abstract}

%%%%%%%%% BODY TEXT
\section{Introduction}

Neural networks are subject to 
shortcut learning, namely a tendency to relying on simple solutions to optimization problems, based on spurious correlations between data and ground truth. Shortcut solutions are thus one of the factors that negatively affect generalization and robustness performance of trained models~\cite{Geirhos_2020,wang2023robustness}.
Mitigating shortcut learning was shown to be beneficial for enhancing the generalization performance and robustness of models~\cite{Saranrittichai,9839408}.  By enforcing models to learn from deeper task-related semantics instead of shallow correlations between data and ground truth that facilitate easy predictions during training, shortcut learning can be effectively addressed~\cite{diagnostics12010040,06922,08822,pezeshki2021gradient,11230}. Existing methods to mitigate the learning of shortcut features include identifying and imitating shortcut features in the other class to reduce their specificity for classification~\cite{diagnostics12010040}, as well as measuring the amount of shortcut information present in the training data~\cite{06922,08822,pezeshki2021gradient,11230}. However, these approaches are often limited to visually observable shortcut features (e.g. color patches and lines) or complex training strategies to learn image representations containing fewer shortcut features.
\begin{figure}
    \centering
    \includegraphics[width = \linewidth]{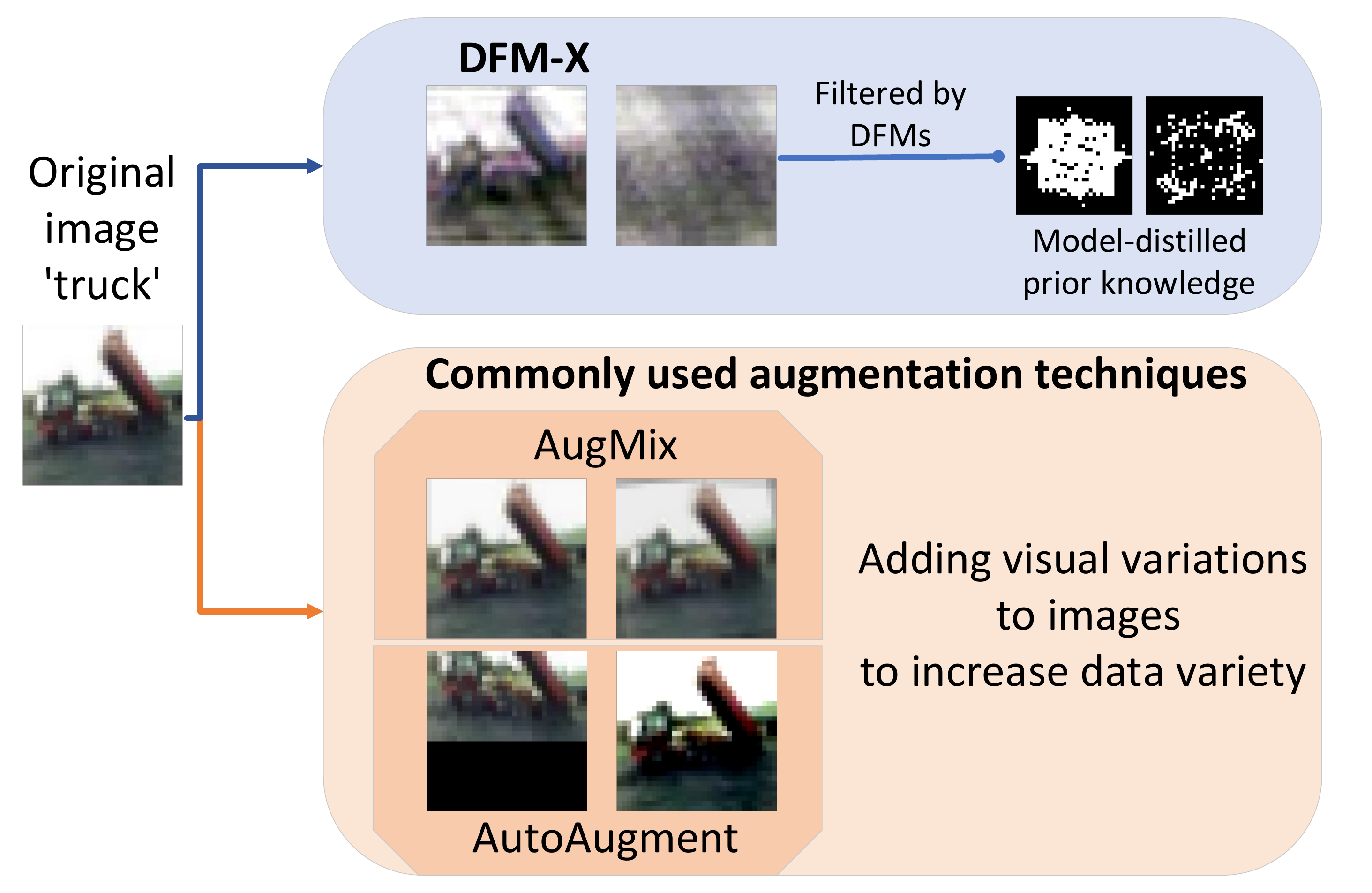}
    \caption{DFM-X exploits the original data efficiently, using model-distilled knowledge about shortcut learning behavior that impairs the generalization and robustness of models, rather than directly adding visual variations to the images like commonly used augmentation techniques.}
    \label{fig:samples}
\end{figure}
Imitating or inducing shortcut features in the images of other classes~\cite{diagnostics12010040}
 is a type of data augmentation. Commonly used data augmentation techniques, e.g. AugMix~\cite{hendrycks2020augmix} and AutoAugment~\cite{cubuk2019autoaugment}, do not usually take shortcut learning into account, but focus more on directly increasing data variety to bridge the distribution gap between training and testing data, improving the generalizability of models.

In this work, we propose a data augmentation method called DFM-X. It is based on prior knowledge about frequency shortcuts~\cite{wang2022frequency,wang2023neural}, which are identified as small sets of specific frequencies that contribute to achieving high-accuracy classification. We compute Dominant Frequency Maps (DFMs) for each class of a previously trained model~\cite{wang2022frequency}, and use them as prior knowledge of where  destructive learning behavior happens in existing models to perform targeted data augmentation.
Our work shares a similar idea with imitating shortcut features, like including color patches specific for a certain class~\cite{diagnostics12010040}, in the images of the other class. In this work, we imitate frequency shortcut features  to reduce the reliance of models on specific frequency sets for classification, thus enforcing models to learn from a wider range of frequencies.
We leverage the algorithm proposed in~\cite{wang2022frequency} to measure the dominant frequencies that play a crucial role in classifying each class, resulting in dominant frequency maps (DFMs). %The DFMs contain frequency shortcuts information, i.e. the frequency sets specific for the classification of certain classes.
% By evaluating the specificity of dominant frequencies to classification, i.e. how well the model performs when provided images with only the dominant frequencies, one can identify frequency shortcuts learned by the model.
We utilize DFMs in our augmentation approach as prior knowledge  (distilled by models from the data) to avoid unwanted learning behavior. This improves the generalizability and robustness of models to common corruptions and adversarial attacks in computer vision.   We demonstrate the difference between DFM-X and other commonly used augmentation techniques in~\cref{fig:samples}.
Compared with AugMix and AutoAugment, DFM-X makes an effort to exploit the original data in an efficient way, using the model-distilled knowledge about learning behavior that impairs the generalization and robustness of models, rather than directly adding variations to the images.  
% The DFMs are used as masks to filter training images. The dominant frequencies of one of the other classes are retained in images of a certain class. Consequently, during training, the dominant frequencies become less specific to their corresponding class, because the  models are forced to utilize these frequencies for the classification of other classes. This compels the models to learn from a wider range of frequencies, and thus more deep and task-related semantics.
%
Our contributions are:
\begin{itemize}
    \item We propose a novel augmentation method called DFM-X to improve the generalization and robustness of models against common corruptions and adversarial attacks without sacrificing their performance on the original test images.
    \item DFM-X exploits model-distilled prior knowledge from data about frequency shortcuts, targeting the mitigation of destructive learning behavior which impairs generalization, unlike commonly used augmentation techniques that focus on increasing data variety directly but rarely consider implicit problems in the data.
    % \item Our approach targets the shortcuts that naturally exist in the data, unlike existing works that focus on those added artificially, e.g. color patches~\cite{diagnostics12010040} and watermarks~\cite{08822}. 
    % \item DFM-X can be integrated with  other augmentation techniques like AugMix~\cite{hendrycks2020augmix} and AutoAugment~\cite{cubuk2019autoaugment} to further enhance the robustness of models. 
\end{itemize}
% Our results show that DFM-X can improve the robustness of models against common corruptions, and it can be easily incorporated with other augmentation techniques, which further improves corruption robustness.  

\section{Related works}
We review existing research related to shortcut learning mitigation and data augmentation in the frequency domain.
\paragraph{Shortcut learning mitigation.}
Avoiding learning shortcuts in the data is a promising approach to improve generalization and corruption robustness by encouraging models to learn more meaningful task-related semantics. Existing work has explored different strategies to address shortcut learning and its impact on model performance.

One approach is to explicitly identify and manipulate shortcuts present in the data. The authors in~\cite{diagnostics12010040} identified shortcuts (i.e. color patches) in a class and induced similar patches in the other class. This forces the models to ignore spurious correlations between the color patches and the class, thus effectively mitigating the influence of shortcuts. 
% However, identifying and manipulating imperceptible shortcuts, e.g. frequency shortcuts~\cite{wang2022frequency} that are specific sets of frequencies corresponding to textures or shapes in the spatial domain, remains a challenge.
Another line of research focuses on addressing shortcut learning without explicitly identifying shortcuts. The work of~\cite{pezeshki2021gradient} proposed a regularization term that decouples feature learning dynamics, allowing the networks to learn from as many features as possible rather than a subset of features that easily minimizes cross-entropy loss. Similarly, \cite{06406} used an auxiliary network with low capacity to measure the degree of shortcut information in images, because image classes containing shortcuts  are easier to learn in early training stages and a low-capacity network is more prone to shortcut learning than a high-capacity one. Leveraging this, the target network with high capacity can selectively learn less from images with high shortcut degrees. Other methods use gradient-based scores~\cite{ahn2022mitigating} to measure the shortcut degree of training samples or adversarial training~\cite{08822,11230} to learn image representations containing less shortcut information.
% Implicit Feature Modification (IFM) introduced by~\cite{11230} tackles shortcut learning by adding adversarial perturbations to representations, enabling models to learn from more predictive features without suppressing other useful features. However, for simple tasks like  color classification, the extracted features may become redundant and less effective.

Existing methods mainly focus on mitigating learning shortcut features that are visually observable.
% Existing methods focus primarily on improving the generalization performance of models on unseen datasets. The potential of shortcut learning mitigation in enhancing the robustness of models against common image corruptions and adversarial attacks is relatively overlooked. 
Our work, instead, aims at the mitigation of shortcut implicit in the data from a frequency perspective. We exploit the learned frequency shortcuts as prior knowledge of unwanted learning behaviors, and learn to avoid them by using the proposed DFM-X augmentation strategy in the training. 

% can benefit corruption robustness and adversarial robustness, filling this gap in the literature.

% However, the feasibility of these approaches in addressing implicit shortcuts present in the data requires further investigation.

\paragraph{Frequency-based data augmentation.}
Data augmentations applied to images are usually spatial transformations, such as flipping, rotation, and cropping. These are commonly used in augmentation techniques such as AugMix~\cite{hendrycks2020augmix}, AutoAugment~\cite{cubuk2019autoaugment}, AugMax~\cite{wang2022augmax}, among others. 
Inspired by the research analyzing the learning behavior in the frequency domain of neural networks (NNs), there is a trend in developing frequency-based augmentation techniques. Chen~\emph{et al.}~\cite{Chen_2021_ICCV} discovered that enforcing NNs to learn more from the phase spectrum than the amplitude spectrum can improve model robustness toward common corruptions. Xu~\emph{el al.}~\cite{Xu_2021_CVPR} proposed amplitude-mixed augmentation, where NNs are trained with phase-invariant images with fused amplitude spectrum because the phase information is considered to be robust to domain change. 

Rather than mixing frequency information, the work of~\cite{8803787} drops frequency components of images if their discrete cosine transformation coefficients are below a randomly selected threshold.   Inspired by the work~\cite{NEURIPS2019_b05b57f6} which demonstrates how noise consisting of different frequency affect classification performance, Soklaski~\emph{et al.}~\cite{soklaski2021fourierbased} added Fourier-basis noise to the operation candidate pool in the AugMix framework~\cite{hendrycks2020augmix}.  
The work in~\cite{10077598,Yang_2020_CVPR} proposed Fourier domain adaptation for segmentation tasks and deep metric learning respectively, which replaces the low frequency of target images with that of source images. As low frequency contains shape information, the annotations of the source images are used as ground truth for training. 

\begin{figure*}[t!]
    \centering
    \includegraphics[width = 0.72\linewidth]{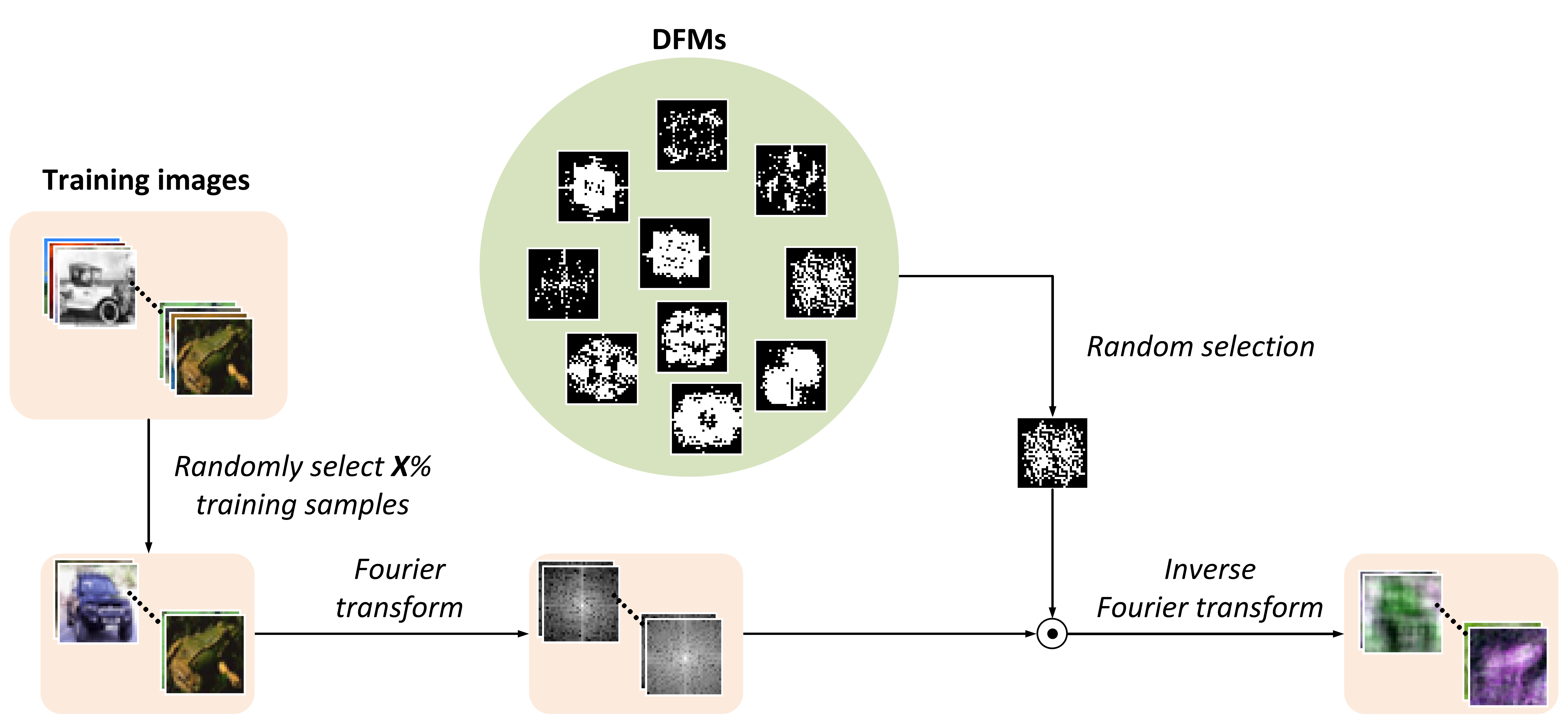}
    \caption{The scheme of DFM-X augmentation. For each training epoch, X\% training images are randomly selected for augmentation. 
    A randomly chosen DFM serves as a mask to filter the Fourier spectrum of the images. If the DFM and the image belong to the same class, the image is not processed. Thus, the images of class $i$ are filtered with the DFM of class $k$ ($i\neq k$).
     This reduces the specificity of the frequency sets to the classification of the corresponding classes, thus mitigating frequency shortcut learning.  }
    \label{fig:scheme}
\end{figure*}

Most augmentation approaches focus on increasing data variety to bridge the distribution gap between training and testing data, or enforcing specific characteristics that benefit the performance. However, they do not consider shortcuts in the data during training. We develop an augmentation strategy based on  prior knowledge about frequency shortcuts that we gain by analyzing models trained for image classification. We devise a form of  augmentation, in which the models are induced to exploit a larger amount of frequency components and avoid learning shortcut solutions,  thus improving model robustness against common corruptions and adversarial attacks.

\section{Methods}
As discovered in~\cite{wang2022frequency}, convolutional neural networks can use small, specific sets of frequencies, i.e. frequency shortcuts, to classify images of certain classes. Because shortcuts harm the generalization of models, we aim to develop an augmentation technique, to improve model robustness and generalization performance by mitigating the learning of frequency shortcuts (prior knowledge distilled from data). We achieve this by reducing the reliance of models on specific frequency sets for the classification of shortcut-affected classes. The models thus rely on a wider range of frequencies to classify images and are induced to learn more semantics. We further evaluate the benefits on corruption robustness and adversarial robustness of models. 

\subsection{DFM-X augmentation}
CNNs can be biased toward specific sets of frequencies to achieve classification~\cite{wang2022frequency}. Our goal is to reduce such bias and enforce the models to learn more semantics, by inducing the use of larger sets of frequencies. Hereby, we design a DFM-based  augmentation technique.

\paragraph{Obtaining DFMs: model-distilled prior knowledge.}
DFMs record the importance of each frequency to the classification of a certain class. 
They can carry knowledge of shortcuts in the data which are learned by a model. We use them as priors in our augmentation approach to avoid unwanted shortcut learning behavior, exploiting data more efficiently and resulting in robust models. 

The algorithm in~\cite{wang2022frequency} computes DFMs by evaluating the importance of frequencies from images based on the degradation of classification performance. To compute the DFM of a certain class, they iteratively remove an individual frequency from the Fourier spectrum of images of the class in the test dataset, and measure the loss in classification. 
If the degradation is above a certain threshold, the frequency is considered important to the classification of the class and it is kept in the Fourier spectrum of images for the following iterations. Otherwise, less important frequencies are removed. We limit the performance degradation to be within 30\%  when the models classify the images of the class retaining only the dominant frequencies, compared to the standard performance.
One can obtain DFMs after the importance of each frequency of the images of the corresponding classes is measured,  which are in the form of binary masks demonstrating the specific sets of frequencies possibly used as shortcuts for classification (see examples of DFMs in~\cref{fig:scheme}). 
% If a model can achieve high-accuracy classification for a certain class on the images containing  the set of frequencies only, it is considered to use a frequency shortcut for classification.
Through leveraging the information contained in the DFMs, we guide the learning behavior of models, aiming to reduce their reliance on specific sets of frequencies associated with shortcut learning.

\paragraph{Augmentation strategy.}
We show the schematic of our augmentation strategy in~\cref{fig:scheme}.
Given a dataset containing images $\{x_m^c\}$ where $c$ is the class of the $m^{th}$ image in the dataset, we compute the DFM of each class for a model $f$. We use the DFMs as priors to guide the training of new models, as they can carry information about unwanted learning behavior.
We randomly select X\% training images to be augmented.  This helps to control the impact of augmentations by adjusting the number of images being augmented.  The selected images are augmented by retaining the dominant frequencies of other classes. That is, we use the DFMs as masks to filter the Fourier spectrum of the images:
\begin{equation*}
   \hat{x}_m^{i} =\mathcal{F}^{-1}[\mathcal{F}[ x_m^{i}] \odot DFM^{k}]\ \ (i \neq k),
\end{equation*}
% we train another model with the same architecture $f'$. 
where $ x_m^{i}$ is the $m^{th}$ image in the dataset of class $i$, $DFM^{k}$ is the dominant frequency map of a randomly selected class $k$, $\mathcal{F}$ and $\mathcal{F}^{-1}$ indicate the Fourier transform and the inverse transform, and $\hat{x}_m^{i}$ is the augmented version being filtered by the DFM. We apply element-wise multiplication of the Fourier spectrum of $x_m^{i}$ and $DFM^{k}$ (the process of filtering), and obtain the augmented image $\hat{x}_m^{i}$ through computing the inverse Fourier transform of the filtered Fourier spectrum of $x_m^{i}$.  
Note that, $i$ is not equal to $k$. This enforces that the models learn from the dominant frequencies of class $k$ to classify the images of class $i$. To sum up, DFM-X augmentation mitigates frequency shortcut learning by highlighting features or visual cues across the whole dataset that are originally specific for certain classes.

\subsection{Evaluation of corruption robustness}
Common image corruptions are visual transformations applied to images and might affect the ability of models to extract semantic features (e.g. Gaussian noise and defocus blur~\cite{hendrycks2019robustness}), thus negatively influencing model robustness. 
We utilize the mean corruption error ($\mathrm{mCE}$) and the relative corruption error ($\mathrm{rCE}$) to evaluate the corruption robustness of models on datasets containing sub-datasets, e.g. CIFAR-C that are corrupted by one corruption~\cite{hendrycks2019robustness}, computed as:
    \begin{equation*}
    \label{equ:1}
    \mathrm{mCE} = \frac{1}{|C|}\sum_{c \in C }\frac{\sum_{s=1}^5 \mathrm{CE}_{s,c}^f}{\sum_{s=1}^5 \mathrm{CE}_{s,c}^{baseline}},
    \end{equation*}
%\noindent and 
\begin{equation*}
    \label{equ:rmce}
    \mathrm{rCE} = \frac{1}{|C|}\sum_{c\in C}\frac{\sum_{s=1}^5 (E_{s,c}^f-E_{clean}^f)}{\sum_{s=1}^5 (E_{s,c}^{baseline}-E_{clean}^{baseline})}, 
    \end{equation*}
\noindent where $\mathrm{CE}_{s,c}^f$ is the classification error of model $f$ on a test set corrupted by $c$ (e.g. defocus blur and shot noise) with severity $s\in \{1,2,3,4,5\}$. The higher the severity, the more influence the corruption effect has on the images.  $C$ is the set of corruptions in the entire test set and $baseline$ is the baseline model for comparison.
The $\mathrm{mCE}$ measures the relative classification performance of a model normalized by that of the baseline. The $\mathrm{rCE}$ additionally measures the performance degradation of model $f$ on corrupted images w.r.t. their clean version. When $\mathrm{mCE}$ and $\mathrm{rCE}$ are less than one, this indicates that model $f$ is more robust than the baseline, as it has less classification error in general. Additionally, we use standard accuracy (SA) to evaluate the performance of models on the original test dataset. The robust accuracy (RA), instead, is the average accuracy of the models tested on the corrupted versions of the test set. 
\subsection{Evaluation of adversarial robustness}
We evaluate the accuracy of models under FGSM and PGD attacks.  These attacks usually have a bias toward high-frequency. As shown in~\cite{chan2022does}, inducing low-frequency bias to models during training can improve adversarial robustness. Differently, our augmentation approach enforces models to learn from a wider range of frequencies with the model-distilled prior knowledge from data, which might be beneficial for adversarial robustness.
% The proposed method avoids unwanted learning behavior instead of aligning the frequency bias of models and data. 
We use $L_{\infty}$-norm bounded perturbation $\mathbf{\epsilon}$ ranging from $1/255$ to $10/255$. For the PGD attack, we use $10$ steps and set the step size $2.5 \epsilon /10 $ to ensure that the boundary of the $\epsilon$-ball is reached. 

\section{Experiments and results}
\subsection{Datasets}
We use CIFAR-10~\cite{Krizhevsky_2009}, which contains 10 classes of 50000 training images and 10000 testing images. For the evaluation of corruption robustness, we use its corrupted variant CIFAR-C~\cite{hendrycks2019robustness}, which includes 19 corrupted subsets. The 19 corruptions are categorized into four groups, including noise (Gaussian, impulse, shot, speckle), blur (defocus, glass, Gaussian, motion, zoom), weather (brightness, fog, frost, snow, spatter), and digital transformation (contrast, elastic, JPEG compression, pixelate, saturate). For each corruption, there are five levels of severity. High severity indicates a high impact of corruption on images.

\subsection{Training setup}
\begin{figure}
    \centering
    \includegraphics[width = \linewidth]{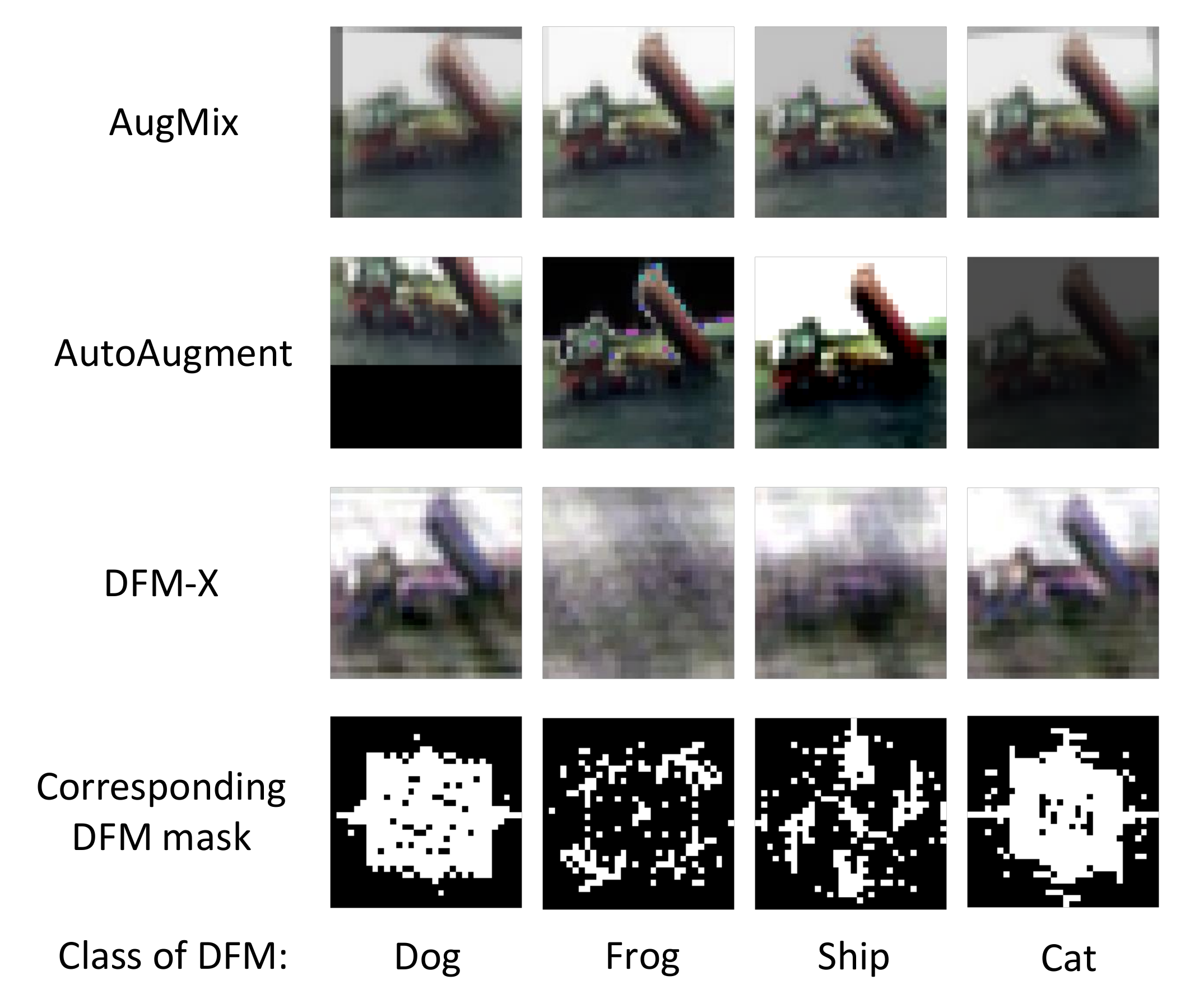}
    \caption{An image of truck is augmented by AugMix (the first row), AutoAugment (the second row), and DFM-X (the third row). The fourth row demonstrates the corresponding DFM used to obtain the images in the third row. }
    \label{fig:samples_aug}
\end{figure}

We train ResNets for 200 epochs on the CIFAR-10 dataset. The initial learning rate is $0.01$, reduced by a factor of 10 if the validation loss does not decrease for 10 epochs. We use  batch size 64 and an SGD optimizer with momentum 0.9 and weight decay $10^{-4}$. Note that, low-capacity models are more prone to shortcut learning than high-capacity models~\cite{06406} and shortcuts in the data are supposed to be architecture-agnostic. Thus, we compute the DFMs of ResNet18, a relatively low-capacity model, in DFM-X augmentation. To compare DFM-X with other commonly used augmentation techniques, we train models with AugMix or AutoAugment. Example images augmented by AugMix, AutoAugment and the proposed DFM-X are shown in~\cref{fig:samples_aug}.

\begin{table}
    \centering
    \scriptsize
    \begin{tabular}{lcccc}
        \toprule
       \bfseries Model  &\bfseries SA  &\bfseries RA &\bfseries mCE (\%) &\bfseries rCE (\%) \\
        \midrule
        ResNet18  &  92.15  & 77.49       & 100 & 100 \\%
        + DFM-30      & 92.11     & 80.43  & 87.94 & \textbf{75.92} \\%
        + DFM-50      &  92.10     & \textbf{81.4}  & 86.61 & 78.73 \\%
        + DFM-70      &   \textbf{92.27}  & 81.2  &\textbf{ 85.36} & 76.74 \\%
        \cmidrule{2-5}
        + AugMix    &  \textbf{93.39}  & 83.47 & \textbf{76.42} & 72.71 \\%
        + AugMix + DFM-30     &   91.44  & 83.65 & 80.68 & {58.18} \\%
        + AugMix + DFM-50     &  91.99   & \underline{\textbf{84.56}} & {77.33}& 62.58\\%
        + AugMix + DFM-70     &   91.08  & {84.2} &80.3 &\underline{\textbf{57.19}} \\%
        \cmidrule{2-5}
        + AutoAugment    &  \underline{\textbf{93.47}}  & 82.78 & 75.81 & 65.86 \\%
        + AutoAugment + DFM-30     &  {92.43}   & 83.29 & {78.28}& 64.72 \\%
        + AutoAugment + DFM-50     &   {92.72}  & \textbf{84.37} & \underline{\textbf{73.23}} & 59.48 \\%
        + AutoAugment + DFM-70     &  91.39   &83.65  & 79.97 & \textbf{59.45}\\%
        \midrule
         ResNet34  & 93.02 & 79.84       & 90.16 & 93.34 \\%
        + DFM-30      & \textbf{93.75} & \textbf{82.32}       & \textbf{78.65} &80.76 \\%
        + DFM-50      & 93.51 & 81.57       & 80.59 & 79.07\\%
        + DFM-70      & 92.7 & 81.93       & 81.7 & \textbf{72.26} \\%
        \cmidrule{2-5}
        + AugMix    &92.18  &    83.71    & 79.69 & 66.63\\%
        + AugMix + DFM-30     & \textbf{93.49}  &    \underline{\textbf{86.42}}    & \underline{\textbf{65.81}} & \underline{\textbf{54.1}}\\%
        + AugMix + DFM-50     & 92.74  & 85.77       & 70.06 & \underline{\textbf{54.1}}\\%
        + AugMix + DFM-70     &  93.28 &  85.45      & 69.27 & 58.14 \\%
        \cmidrule{2-5}
        + AutoAugment    & \underline{\textbf{94.08}} &    83.66    & 72.97 & 70.32 \\%
        + AutoAugment + DFM-30     & 93.71 & 85.72  & \textbf{66.38} & 57.58 \\%
        + AutoAugment + DFM-50     &  93.46  & 85.55       & 67.42 & \textbf{56.36} \\%
        + AutoAugment + DFM-70     & 93.35  &   \textbf{85.96}     & 67.53 & 57.06 \\%
        % \midrule
        %  ResNet50  & \textbf{91.95} &  77.47      & 99.28 & 92.73 \\%
        % +DFM-30      & 91.68  &  \textbf{80.64}  & \textbf{92.23} & 85.62 \\%
        % +DFM-50      & 91.65 &  79.33 & 93.97 & 83.94 \\%
        % +DFM-70      & 91.27 & 80.32  & 92.79 & \textbf{80.2} \\ %
        % \cmidrule{2-5}
        % +AugMix    & 92.5 &  82.93      & 82.19 & 74\\%
        % +AugMix+DFM-30     & \textbf{92.79}  &   85.37     & 73.22 & 62.93\\
        % +AugMix+DFM-50     & 92.58 &  \textbf{85.82}      & \textbf{71.31} & \textbf{56.45}\\
        % +AugMix+DFM-70     & 92.21  &    84.86    & 75.13 & 59.22\\
        % \cmidrule{2-5}
        % +AutoAugment    & 93.79 &    82.37    & 77.63 & 75.79\\%
        % +AutoAug+DFM-30     &  \textbf{93.87}  &    84.48   & 70.45  & 65.8 \\%
        % +AutoAug+DFM-50     &  93.56  &    \textbf{84.65}    & \textbf{70.16} &  62.04\\%
        % +AutoAug+DFM-70     &  92.36  &   84.06   & 75.6 & \textbf{60.18} \\%
        \bottomrule
    \end{tabular}
    \caption{Performance of ResNets on CIFAR-10 and CIFAR-C. The best values of each group of models are in bold and the best values for ResNet18 and ResNet34 are underlined.}
    \label{tab:result_cifar}
\end{table}

\subsection{Robustness against common corruption}
We report the results of the models trained with different augmentation strategies in~\cref{tab:result_cifar}. The best values of models with the same architecture trained with namely  DFM-X, AugMix + DFM-X and AutoAugment + DFM-X, are highlighted in bold, and the best values for ResNet18 and ResNet34 are underlined. 
\paragraph{DFM-X benefits corruption robustness.}
ResNets trained with DFM-X augmentation are more robust against common corruptions than ResNets trained without DFM-X. They have higher or comparable standard accuracy than ResNets trained without DFM-X, as well as higher robust accuracy. This indicates that DFM-X benefits the robustness of models to common corruptions without impairing their performance on the clean dataset. 
% The models have low mCEs (around 86\%) and rCEs (around 76\%). 
We conjecture that DFM-X enforces models to learn from a wider range of frequencies with the prior knowledge provided, and thus more meaningful and task-related semantics is used by the models, benefiting their corruption robustness.
% Selecting different X\% of images for augmentation does not significantly affect the corruption robustness. 

\paragraph{Comparison with existing augmentations.}
We compare DFM-X to existing and largely-used augmentation techniques like AugMix~\cite{hendrycks2020augmix} and AutoAugment~\cite{cubuk2019autoaugment}. The models trained with DFM-X, AugMix, and AutoAugment have similar SA and RA, while the model trained with DFM-X has higher mCE than the models trained with AugMix or AutoAugment. The rCE of models trained with DFM-X is also higher than that of models trained with AutoAugment or AugMix. We attribute this to the fact that other augmentation techniques focus on increasing data variety to reduce the distribution gap between training and testing data,  and may use augmentations that are visually similar to the corruptions in CIFAR-C. Our approach, instead, focuses on exploiting as much information as possible from the clean training data without additionally overlaying corruption-like variations, learning more meaningful and task-related semantics. 
% DFM-X mitigates learning shortcuts in the frequency domain, rather than directly targeting generalization to spatial transformations. 
 We thus investigate the effectiveness of combining DFM-X with another augmentation technique, as they augment images differently (DFM-X exploits data efficiently while the others increase data variety by adding corruption-like variations).
 
 % one of them augments images from the frequency perspective, and another augments images in the spatial domain. 

% The rCE of models trained with DFM-X is close to that of ResNet18+AugMix, and is higher than that of ResNet18+AutoAug. From the average accuracy of the models tested on each corruption set in~\cref{fig:avg_acc}, we observe that DFM-70 improves the accuracy in terms of glass blur and pixelate to a large extend, while the improvement to other corruption types is less outstanding. This explains why the accuracy of the three models is close to each other, while there are large deviations in the values of mCE and rCE. 

% \begin{figure}
%     \centering
%     \includegraphics[width = 0.9\linewidth]{figures/Picture2.png}
%     \caption{Average accuracy of each corruption for ResNet18+DFM-70/AutoAug/AugMix. DFM-70 improves the accuracy in terms of glass blur and pixelate significantly, compared to AutoAug and AugMix.}
%     \label{fig:avg_acc}
% \end{figure}

\paragraph{Boosted robustness with a complementary technique.}
We apply DFM-X together with either AugMix or AutoAugment during training, and observe that this contributes to further improving corruption robustness (\cref{tab:result_cifar}).
For example, ResNet18 trained with AugMix/AutoAugment and DFM-50 augmentations generally achieve higher robust accuracy than the models trained with only AugMix or AutoAugment. Moreover, they have lower or comparable values of $\mathrm{mCE}$ and $\mathrm{rCE}$, compared to those trained with AugMix or AutoAugment only.
When ResNet34 is trained with DFM-30 and AugMix or AutoAugment augmentations, the models demonstrate more robustness than those trained with only one of the augmentation techniques.  This indicates that combining augmentation techniques that are complementary can further benefit corruption robustness. For the future design of augmentation techniques, we should focus on inspecting data itself and exploiting it more efficiently,
rather than directly increasing the variety of data by adding visual variations. 

Intriguingly, when ResNet18 and ResNet34 are trained with AugMix, both models show similar corruption robustness, though ResNet34 has a larger model capacity than ResNet18. 
% The values of robust accuracy are around 83.5\% and 
ResNet34 + AugMix even has a slightly worse $\mathrm{mCE}$ than ResNet18 + AugMix. However, when models are trained with AugMix + DFM-X, ResNet18s have worse  $\mathrm{mCEs}$ than that trained solely with AugMix.
ResNet34 trained with AugMix + DFM-30, instead, demonstrates significantly improved corruption robustness (mCE equal to $65.81$). As low-capacity models are more prone to shortcut learning than high-capacity models, DFM-X together with AugMix might not be enough to mitigate shortcut learning in ResNet18 but benefits the robustness of ResNet34.  
We conjecture that for low-capacity models, there needs extra regularization to overcome shortcuts. 
% might not be able to distinguish the difference between the displacement effect and the shortcut patterns, thus it fails in mitigating shortcut learning. 
Moreover, AugMix results in displacement effects that are visually similar to those in some DFM-augmented images (see~\cref{fig:samples_aug}). Thus, there is an partial overlap in the augmentation effects.
When combining DFM-X with other augmentation techniques, it is important to consider what kinds of operation are appropriate and complementary to DFM-X. 

\paragraph{The choice of X.}
Our results show that the percentage of training images subject to augmentation via DFM-X does not influence significantly on the corruption robustness when the models are trained with DFM-X solely (they have close RA, mCE and rCE). However, when DFM-X is incorporated with AugMix or AutoAugment, models with a different capacity might perform better as a different value of X is chosen. For instance, ResNet18 prefers DFM-50 while ResNet34 prefers DFM-30. We attribute this to model capacity. As a relatively low-capacity model, ResNet18 suffers more from shortcut learning than ResNet34, and thus it needs more regularization during training. The more images are augmented, the more regularization is imposed on the training process. 
 
% Models trained with DFM-50 and DFM-80 also gain more robustness when another augmentaion is additionally used in training. Among the models trained with AugMix, DFM-50 further improves the robust accuracy while DFM-70 reduces rCE to a large amount. 

% For the models trained with AutoAugment, ResNet18+AutoAug+DFM50 achieves the highest robust accuracy and the best mCE. Its rCE is only 0.03 lower than that of the one trained with DFM-70, which has the best rCE 59.45. Compared to the models in other groups, it performs evenly in terms of the robust accuracy, mCE, and rCE. Its robust accuracy (84.37) is close to the highest value (84.56) among all the models while the rCE is only 2.26 higher than the lowest rCE of ResNet+AugMix+DFM-70. Moreover, it has the best mCE compared to all the other models. The results demonstrate that DFM-X can be easily adapt to other augmentation techniques and further improve corruption robustness.    

% Models trained AutoAugment and DFM-X in general perform better that those trained with AugMix and DFM-X. This might due to the combined augmentations used in AugMix result in a displacement effect that is visually similar to  those in the DFM-augmented images. Thus, DFM-X combining AugMix does not gain more robustness compared to AutoAugment. 

\begin{table}
    \centering
    \scriptsize   
    \begin{tabular}{lcccc}
        \toprule
      \bfseries  Model  &\bfseries Noise &\bfseries Blur &\bfseries Weather &\bfseries Digital  \\
        \midrule
        ResNet18 & 100 & 100  &  100  & 100 \\
        + DFM-30      & 88.75    & 91.6  & 85.2  & 86.6\\
         + DFM-50       &   \textbf{78}  & 94.6  & \textbf{84.8}  &87.2\\
         + DFM-70       &   86.5  & \textbf{91}  & 84  &\textbf{79.7}\\
        \cmidrule{2-5}
         + AugMix      &  62   &\textbf{ 86.2 }& \textbf{75.2} &\textbf{79.8} \\
         + AugMix + DFM-30      &  {53.5}   & 92.2  &  85.6 &86.2\\
         + AugMix + DFM-50      &  {50.75}   &  {91} & 79.6  &82.6\\
         + AugMix + DFM-70      &   {\textbf{50.5}}  & 95.4  &  83.8 &85.8\\
        \cmidrule{2-5}
         + AutoAugment      &   72.75  & {\textbf{85.4 }}&{\textbf{ 63.8}} &80.8\\
         + AutoAugment + DFM-30      &   62  & 95.8  & {73.4}  &{78.2}\\
         + AutoAugment + DFM-50      &   61.75  &  {89} & {68.6}  &{\textbf{71}}\\
         + AutoAugment + DFM-70      &  \textbf{61.5}   & 97.8  & {77.4}  &{79.8}\\
        \bottomrule
    \end{tabular}
    \caption{The corruption error (CE) (\%) of ResNet18s trained with different augmentation techniques on each corruption type (Baseline: ResNet18). The best values in each group are highlighted in bold.}
    \label{tab:result_cifar_detailed}
\end{table}

 \begin{table*}
    \centering
    \scriptsize   
    \begin{tabular}{lcccccccccc}
        \toprule
        \multirow{2}{*}{\bfseries Model }  &  \multicolumn{10}{c}{$L_{\infty}$-norm bounded perturbation of size $\mathbf{\epsilon}$}     \\
          &  1/255 &  2/255 &  3/255 &  4/255  &  5/255 &  6/255 &   7/255  &  8/255  &   9/255  &   10/255 \\
        
        \midrule
        ResNet18  &  86.47 & 75.66 & 64.03  & 54.3  & 46.46  & 40.05  & 35.04  & 31.83  & 28.92 & 26.47\\
        + DFM-50   & {87.71} & {78.97} & {69.43} & {60.7}  & {53.05}  & {46.55} & {41.04}  &  {37.18} & {34.09} & {31.61}\\ % better  
        % + DFM-70  &  87.26 & 78.13 & 69.02 & 60.22  & 52.8  & \textbf{46.79} & \textbf{41.74}  &  \textbf{38.22} & \textbf{35.05} & \textbf{32.73} \\ % better
        \cmidrule{2-11}
         + AugMix  & 88.02 & 79.27 & 69.28 & 60.38 & 52.52 &46.03  & 41.17 & 36.93 & 33.57 & 30.69\\
        + AugMix + DFM-50 & \textbf{88.29} & \textbf{79.97} & \textbf{71.07} &\textbf{ 62.41} &\textbf{ 55.23} & \textbf{48.75} & 43.38 & 38.92 & 35.03 & 31.67 \\
        % + AugMix + DFM-70&  &  &  &  &  &  &  &  &  &  \\
         \cmidrule{2-11}
        + AutoAugment  & 87.72 & 76.13 & 65.26 & 55.7 & 48.99 & 43.85 & 39.91 &  36.64& 33.97 & 32.15\\
        + AutoAug + DFM-50  & 87.7 &78.51 &69.03 & 60.68& 54.03&48.33 &\textbf{43.94} & \textbf{40.66}   & \textbf{37.8}   & \textbf{35.43}  \\
        % + AutoAugment + DFM-70  & 87.2 & \underline{\textbf{79.26}} & \underline{\textbf{70.35}} &\underline{\textbf{62.35}} & \underline{\textbf{55.61}} & \underline{\textbf{49.72}} & \underline{\textbf{44.53}} & 40.25 & 36.52 & 33.88 \\
        \midrule
        ResNet34       & 87.7 &  77.53 & 67.61   & 58.12  & 50.71 & 45.21  &  40.32  & 36.69  & 33.52 & 30.91\\
        + DFM-30  & 88.45  & 79.17 & 70.48 & 63.14  & 57.32  & 52.96 & 48.88  & 46.03  & 43.72 & 41.81   \\ %super
        % + DFM-50  &  88.21  & 78.9 & 69.93 & 61.26  &  54.64 & 49.67  & \textbf{45.8}  & \textbf{43.03}  & \textbf{40.49} & \textbf{38.58}   \\ % super
        %  + DFM-70 &  \textbf{88.63} & \textbf{80.68} & \textbf{71.77} & \textbf{63.01}  & \textbf{55.98}  & \textbf{49.88} & 44.91  &  41.44 & 38.51 & 35.98 \\
         \cmidrule{2-11}
         + AugMix & 88.71 & 79.56 & 70.14 & 60.99 & 52.62 & 45.78 & 40.12 & 35.58 & 31.85 & 28.72 \\
        + AugMix + DFM-30 & 88.8 & 81.16 & 72.23 & 63.34 & 55.37 & 48.41 & 42.17 & 37.45 & 33.55 & 30.35 \\
        % + AugMix + DFM-70&  &  &  &  &  &  &  &  &  &  \\
          \cmidrule{2-11}
       + AutoAugment &  87.39 & 76.76 & 65.98 & 57.06 & 50.48 & 45.3 & 41.04 & 38.05 & 35.55 & 33.68  \\
        +AutoAugment + DFM-30  & \textbf{89.23} & \textbf{81.96} & \textbf{74.92} &\textbf{ 68.34} &  \textbf{63.01}& \textbf{58.81} & \textbf{54.93}  & \textbf{51.64} &\textbf{49.13} & \textbf{46.94}\\
        % +AutoAug +DFM-50 &\underline{\textbf{88.92}} & \underline{\textbf{81.69}} &  73.7 & \underline{\textbf{67.07}} & 61.18 &  56.68 & 53.23 & 50.12 &47.43   &  45.16\\
        %  + AutoAug + DFM-70   & 88.66 & 81.45 & \underline{\textbf{73.77}} & 66.97 & \underline{\textbf{61.53}} & \underline{\textbf{57.4}} & \underline{\textbf{53.61}} & \underline{\textbf{50.37}} & \underline{\textbf{47.83}} & \underline{\textbf{45.48}}\\
        \bottomrule
    \end{tabular}
    \caption{Accuracy (\%) of models under FGSM attack. The best values of ResNet18 and ResNet34 are highlighted in bold.}
    \label{tab:result_cifar_fgsm}
\end{table*}

\begin{table*}
    \centering
    \scriptsize   
    \begin{tabular}{lccccccccccc}
        \toprule
      \multirow{2}{*}{\bfseries Model }  &  \multicolumn{10}{c}{$ L_{\infty}$-norm bounded perturbation of size $\mathbf{\epsilon}$}     \\
          &  1/255 &  2/255 &  3/255 &  4/255  &  5/255 &  6/255 &   7/255  &  8/255  &   9/255  &   10/255 \\
        \midrule
        ResNet18  & 85.39 &  70.79 & 53.5  & 38.43  & 26.09 & 18.13  & 12.98  & 9.95 & 7.98 & 6.96 \\
        + DFM-50  &  86.93  &   75.4 & {60.63} &  {45.88} & {33.61}  &  {24.8} &  {18.51}  &   {14.15} &  {11.16} &  {9.09} \\% good
        % + DFM-70   & 86.56 &  73.96 & 58.37  & 43.6 & 31.99  & 23.56  & 17.57 & 13.54  & 10.64  & 8.56 \\% good
         \cmidrule{2-11}
        + AugMix   & 87.36 & 74.91 & 58.75 & 43.35 & 30.54 & 21.49 & 15.52 & 11.57 & 9.07 & 7.41\\
        + AugMix +DFM-50 & \textbf{87.71} &  \textbf{76.86}& \textbf{63.81} & \textbf{51.01} & \textbf{38.22} &\textbf{29.99}  & \textbf{22.95} & \textbf{17.3} &\textbf{13.63}  & \textbf{11.14} \\
        % + AugMix + DFM-70&  &  &  &  &  &  &  &  &  &  \\
         \cmidrule{2-11}
        + AutoAugment   & {86.08}  & 67.37 & 46.96 & 32.17 & 21.48 & 15.03 & 11.29 &  8.63 & 7.01 & 6.1\\
        + AutoAugment + DFM-50   & 86.45 & 71.45 & 54.86 & 40.13  & 28.75 & 20.79 & 15.79 & 12.27 & 9.89 & 8.15 \\
        % + AutoAug + DFM-70   & \textbf{86.61} &  75.4}} & \underline{\textbf{61.95}} & \underline{\textbf{48.59}} & \underline{\textbf{37.05}} & \underline{\textbf{28.09}} & \underline{\textbf{21.25}} & \underline{\textbf{16.54}} & \underline{\textbf{13.09}} & \underline{\textbf{10.76}} \\
       
        \midrule
        ResNet34   & 86.85  & 73.35 & 57.26 & 42.31  &  30.72 & 21.72 & 15.42  &  11.55 & 8.74 & 7.38\\
        + DFM-30 & 87.94  & 75.03 & 58.78 & 44.29  & 33.2  & 25.45 &  20.16 &  16.35 & 13.87 & 11.72  \\ % last good
        % + DFM-50  &  86.98 & 73.29 & 55.99 & 40.9  & 29.41  & 21.47 & 15.93  & 12.25  & 9.88 &  8.23  \\
        % + DFM-70  &  \underline{\textbf{88.03}}  & \underline{\textbf{76.8}} & \underline{\textbf{61.94}} & \underline{\textbf{47.09}}  & \underline{\textbf{35.0}}  & \textbf{26.24} & \textbf{19.5}  &  \textbf{15.33} & \textbf{12.53} & \textbf{9.85} \\ %good.
         \cmidrule{2-11}
         + AugMix & 88.18 & 76.39 & 62.5 & 48.74 & 36.88 & 27.59 & 20.52 & 15.6 & 12.17 & 9.8 \\
         + AugMix + DFM-30 & 88.25 & \textbf{78.38} & \textbf{65.2} & \textbf{51.62} & \textbf{39.6} & \textbf{29.55} &\textbf{22.58}  & \textbf{17.35} & 13.69 &10.93  \\
        % + AugMix +DFM-50 &  &  &  &  &  &  &  &  &  &  \\
        % + AugMix + DFM-70&  &  &  &  &  &  &  &  &  &  \\
        \cmidrule{2-11}
        + AutoAugment    & 86.05 & 68.98 & 49.95 & 34.35 & 24.14 & 17.62 &13.11  & 10.42 & 8.69 & 7.39 \\
        + AutoAugment + DFM-30 & \textbf{89.41}& 76.97 & 61.64 & 46.6 & 35.1 & 26.33 & 20.89  & 16.94 & \textbf{14.32} & \textbf{12.35}\\
        % + AutoAug + DFM-50   &\textbf{86.94}  & 73.25 & 57.11 & 42.97 & 31.96 & 23.88 & 18.55 & 14.6 & 11.88& 9.79  \\
        %  + AutoAug + DFM-70   &{86.81}  & \textbf{73.55} & \textbf{58.14} & \textbf{44.99} & \textbf{34.37} & \underline{\textbf{26.32}} & \underline{\textbf{20.56}} &\underline{\textbf{16.57}} & \underline{\textbf{13.44}} &\underline{\textbf{11.45}}\\
        \bottomrule
    \end{tabular}
    \caption{Accuracy (\%) of models under PGD attack.  The best values of ResNet18 and ResNet34 are highlighted in bold.}
    \label{tab:result_cifar_pgd}
\end{table*}

\paragraph{Robustness against different corruption types.}

In~\cref{tab:result_cifar_detailed}, we present a detailed overview of the robustness of ResNet18 trained with different augmentation techniques against the four corruption categories in CIFAR-C, namely noise, blur, weather conditions, and digital transformation. The best results of ResNet18 trained with DFM-X, AugMix + DFM-X and AutoAugment + DFM-X are highlighted in bold respectively. 
Observed from values in bold, models trained with DFM-X demonstrate better robustness against all types of corruption than the models trained without DFM-X. Among the four corruption types, the improvement in the robustness toward blur corruption is relatively lower than that of the other three types. As demonstrated in~\cite{NEURIPS2019_b05b57f6}, blur corruptions, such as defocus blur and Gaussian blur, have energy highly concentrating on middle-high frequencies.  Our augmentation technique enforces models to look into a wider range of frequencies for classification, and thus, the models are relatively less robust to corruptions having a specific energy concentration in the Fourier spectrum than those having a rather even energy distribution over the spectrum, e.g. Gaussian and shot noise. 

% \[TO\ BE\ REVISED\]
% AugMix+DFM-X encourages better robustness against noise corruption, in a compensation of less robustness to the other three types of corruption.   Differently, AutoAugment+DFM-X show better robustness toward noise and digital corruptions, leveraging the robustness against blur and weather corruptions, compared to the other models. As shown in~\cref{fig:avg_acc}, DFM-X benefits specifically the robustness against glass blur and pixelate, while AutoAugment does not. Combining these two augmentation techniques thus can gain more robustness than using only one of them. 
\begin{figure}
    \centering
    \subfloat[]{\includegraphics[width= 0.9\linewidth]{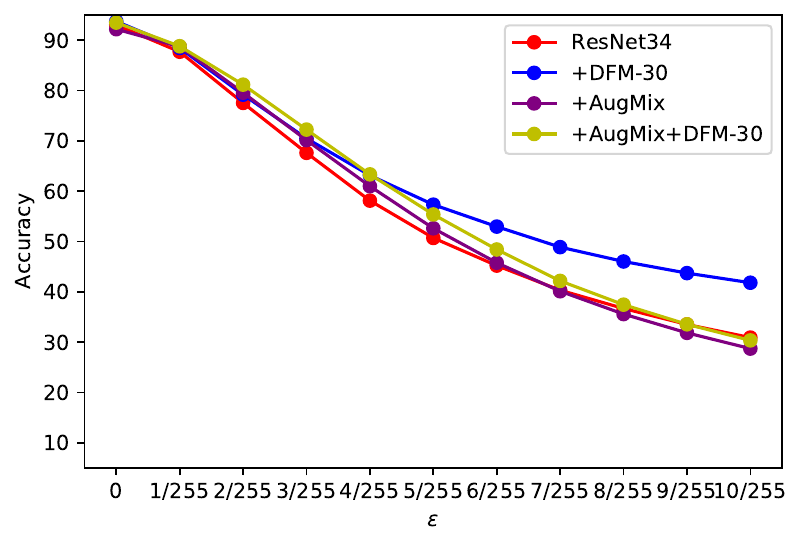}}
    
    \subfloat[]{\includegraphics[width= 0.9\linewidth]{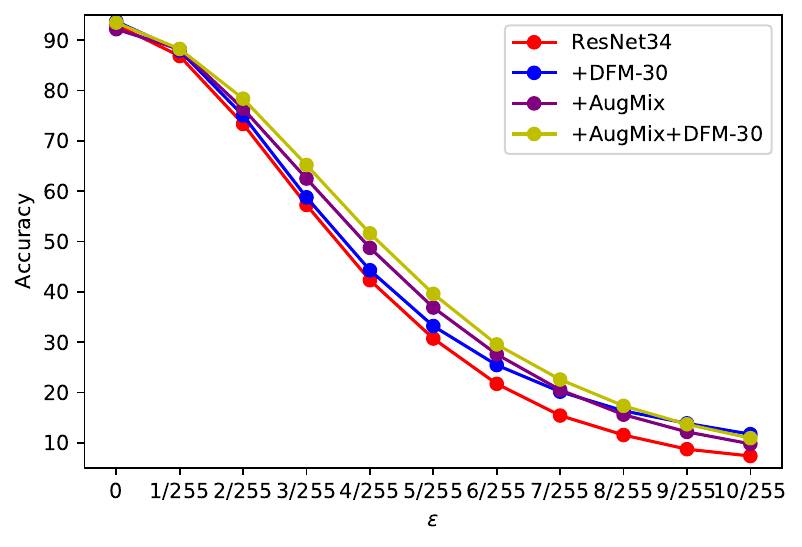}}

    \caption{ResNet34 trained without augmentation or with  DFM-30, AugMix, or AugMix + DFM-30 under (a) FGSM attack and (b) PGD attack. }
    \label{fig:adv_attack_augmix}
\end{figure}

\begin{figure}
    \centering
    \subfloat[]{\includegraphics[width= 0.9\linewidth]{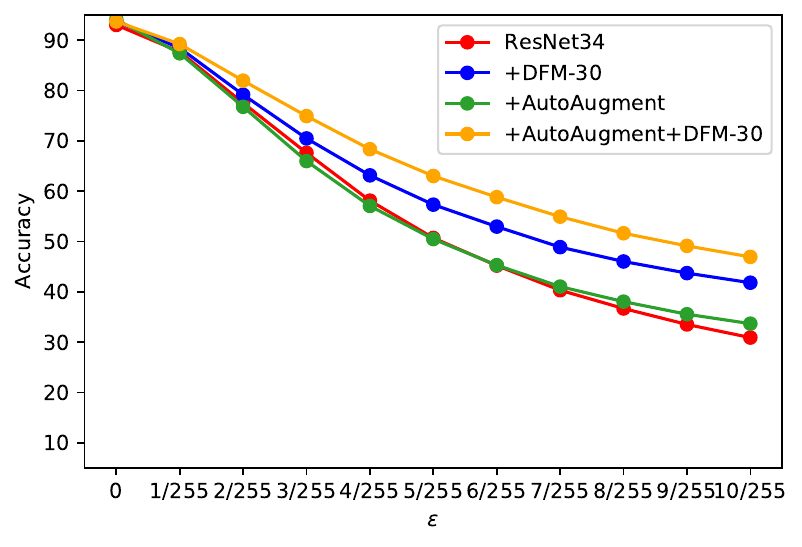}}
    
    \subfloat[]{\includegraphics[width= 0.9\linewidth]{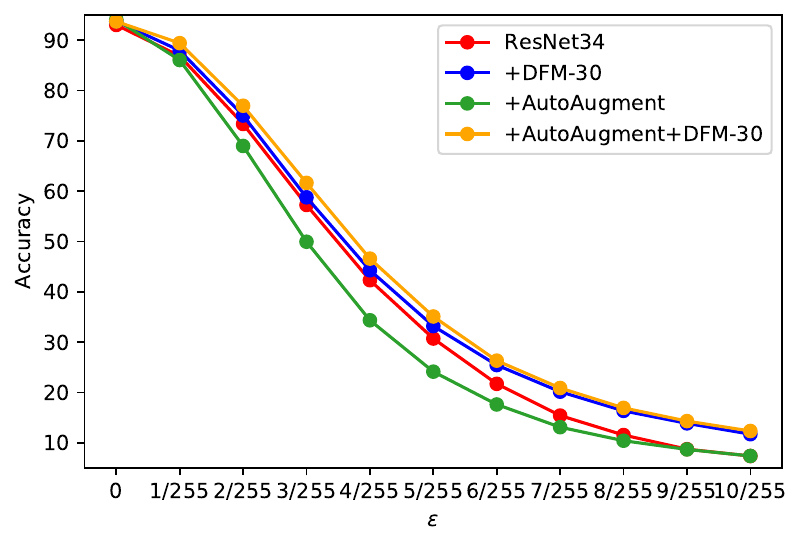}}

    \caption{ResNet34 trained without augmentation or with  DFM-30, AutoAugment, or AutoAugment + DFM-30 under (a) FGSM attack and (b) PGD attack. }
    \label{fig:adv_attack}
\end{figure}

\subsection{Robustness against adversarial attacks}
We evaluate the adversarial robustness of the models trained with different augmentations under FGSM and PGD attacks.
We select ResNet18 + AugMix/AutoAugment + DFM-50 and ResNet34 + AugMix/AutoAugment + DFM-30, as models with a different capacity prefer a different percentage of images to be augmented.
We report the classification accuracy of the models under the FGSM and PGD attacks in~\cref{tab:result_cifar_fgsm,tab:result_cifar_pgd}, respectively. 

\paragraph{Learning from more frequencies improves robustness.}
We observe that ResNets trained with DFM-X show improved adversarial robustness to the FGSM and PGD attacks.  Unlike PGD and FGSM adversarial training which sacrifices  performance on natural images~\cite{madry2019deep}, DFM-X maintains high model performance on the original test set while improving the robustness of models to both FGSM and PGD attacks. We attribute this to the wider learning range of frequencies, when compared to models trained with standard setups.  DFM-X augments images based on the prior knowledge of the reliance of models on specific frequencies, with the aim of reducing it during training. Thus, the frequency bias of models associated with their vulnerability to adversarial noise is reduced, benefiting adversarial robustness. 

\paragraph{DFM-X vs. AugMix.}
In~\cref{fig:adv_attack_augmix} we compare the results of ResNet34 trained without augmentation or with DFM-30, AugMix, and AugMix + DFM-30. We observe that the model trained with AugMix (see the purple line) demonstrates less robustness to FGSM attack, compared to the model trained with DFM-30 (see the blue line). Regarding the PGD attack, both models show comparable robustness. Combining AugMix with DFM-30 does not obtain more robustness to the FGSM attack when $\epsilon$ becomes large, but the model is robust to the PGD attack. From~\cref{tab:result_cifar_fgsm}, ResNet18 trained with AugMix shows similar robustness to the FGSM attack to the one trained with DFM-50, but it is less robust to the PGD attack. 
We conjecture that AugMix, resulting in images with similar displacement effects to those augmented by DFM-X (see~\cref{fig:samples_aug}), might indirectly augment the frequency information of images like DFM-X. But it is less effective than DFM-X, which employs model-distilled prior knowledge from the data in augmentation, rather than randomly adding augmentations to images. 
Combining DFM-X with AugMix avoids unwanted learning behavior and increases data variety, thus obtaining more adversarial robustness.

\paragraph{Reduced negative impact of AutoAugment.}
We demonstrate in~\cref{fig:adv_attack} that ResNet34 trained with DFM-30 (see the blue line) has better adversarial robustness than ResNet34 and ResNet34+AutoAugment. Interestingly, training solely with AutoAugment impairs adversarial robustness slightly to FGSM attack and significantly to PGD attack (see green lines in~\cref{fig:adv_attack}). When combining DFM-30 and AutoAugment (see the orange line), the model gains more robustness to the FGSM attack, compared with the one trained only with DFM-30.
We observe from~\cref{tab:result_cifar_fgsm,tab:result_cifar_pgd}  that AutoAugment also impairs the adversarial robustness of ResNet18 to the FGSM and PGD attacks. Training the model with AutoAugment and DFM-X augmentations benefits the adversarial robustness of the models. 
Although AutoAugment alone harms the robustness, using DFM-X avoids much performance degradation under the attacks. From the augmented images in~\cref{fig:samples_aug}, DFM-X results in different variations from those augmented by AutoAugment. We attribute the models having better adversarial robustness than those trained with only one of them to the complementarity between DFM-X and AutoAugment in terms of augmentation effects.

% DFM-50 benefits the robustness to small perturbations ($\epsilon$ ranges from 1/255 to 5/255) while DFM-70 improves robustness to large perturbations ($\epsilon$ ranges from 6/255 to 10/255).
% This might be because that when the $\epsilon$ increases, the added adversarial noise becomes more apparent than those with small $\epsilon$, resulting in textures that are similar to those appear after DFM filtering. DFM-70 has more images augmented than DFM-50, and thus the model trained with it is more robust to large $\epsilon$. 

 % Combined with AutoAugment with DFM-50/70, both can ease the negative impact of AutoAument on the robustness against FGSM attack to a large extent. Additionally, both models show better adversarial robustness to FGSM attack than those trained only with DFM-X. ResNet34s demonstrate similar performance to ResNet18s. 
 
% As the multi-steps variant of FGSM~\cite{madry2019deep}, ResNet18+DFM-50 performs better than ResNet18 and ResNet18+DFM-70 to PGD attack. But the difference between DFM-50 and DFM-70 is small. Though  AutoAugment has a negative impact on adversarial robustness, DFM-70 combined with AutoAugment demonstrates much better adversarial robustness than other ResNet34 counterparts. ResNet34s perform similarly to ResNet18s in terms of PGD attack. 

\section{Conclusions}
We propose DFM-X, an augmentation approach that leverages prior knowledge about frequency shortcuts. Motivated by shortcut mitigation, our method aims at avoiding unwanted shortcut solutions by enforcing models to learn from a wider range of frequencies and thus more semantics. DFM-X exploits data efficiently, as it targets implicit problems in the data that might impair the generalization and robustness of models, unlike other commonly used augmentation techniques focusing on increasing data variety by adding visual variations.
Our experimental results show that DFM-X enhances model robustness against common corruptions and adversarial attacks without sacrificing the standard performance on the original test set. 
Combining DFM-X with other commonly used augmentation techniques, e.g. AugMix and AutoAugment, gains more robustness than using only one of them. DFM-X compensates for the weakness of AutoAugment in impairing adversarial robustness. We observe that the complementarity between augmentation techniques is important to model performance.  Distilling prior knowledge about destructive learning behavior from data helps exploit data more efficiently. We suggest future research on designing augmentation strategies that consider data characteristics instead of directly increasing the visual variations of images to bridge the distribution gap between training and testing data. 

\noindent\textbf{Acknowledgements. }
This work was supported by the \href{https://sites.google.com/view/search-utwente}{SEARCH project}, UT Theme Call 2020, Faculty of Electrical Engineering, Mathematics and Computer Science, University of Twente.

 {\small
\bibliographystyle{ieee_fullname}
\bibliography{egbib}
}

\end{document}